\def\year{2018}\relax
\documentclass[letterpaper]{article}
\usepackage{aaai18}
\usepackage{times}
\usepackage{helvet}
\usepackage{epsfig}
\usepackage{epstopdf}
\usepackage{mathtools}
\usepackage{amsfonts}
\usepackage{courier}
\usepackage{color}
\usepackage{makecell}
\usepackage{subfig}
\usepackage{diagbox}

\newcommand{\PreserveBackslash}[1]{\let\temp=\\#1\let\\=\temp}

\newcolumntype{C}[1]{>{\PreserveBackslash\centering}p{#1}}
\newcolumntype{R}[1]{>{\PreserveBackslash\raggedleft}p{#1}}
\newcolumntype{L}[1]{>{\PreserveBackslash\raggedright}p{#1}}

\begin{document}
\title{Graph Correspondence Transfer for Person Re-identification}
\author{
Qin Zhou$^{1,3}$ \ Heng Fan$^3$ \ Shibao Zheng$^1$ \ Hang Su$^4$ \ Xinzhe Li$^1$ \ Shuang Wu$^5$ \ Haibin Ling$^{2,3,*}$ \\
$^1$Institute of Image Processing and Network Engineering, Shanghai Jiao Tong University, Shanghai 200240, China\\
$^2$Computer Science and Engineering, South China University of Technology, Guangzhou 510006, China\\
$^3$Department of Computer \& Information Sciences, Temple University, Philadelphia 19122, USA\\
$^4$Department of Computer Science and Technology, Tsinghua University, Beijing 100084, China\\
$^5$YouTu Lad, Tencent, Shanghai 200233, China\\
}
\maketitle
\begin{abstract}
In this paper, we propose a graph correspondence transfer (GCT) approach for person re-identification. Unlike existing methods, the GCT model formulates person re-identification as an {\it off-line} graph matching and {\it on-line} correspondence transferring problem. In specific, during training, the GCT model aims to learn {\it off-line} a set of correspondence templates from positive training pairs with various pose-pair configurations via patch-wise graph matching. During testing, for each pair of test samples, we select a few training pairs with the most similar pose-pair configurations as references, and transfer the correspondences of these references to test pair for feature distance calculation. The matching score is derived by aggregating distances from different references. For each probe image, the gallery image with the highest matching score is the re-identifying result. Compared to existing algorithms, our GCT can handle spatial misalignment caused by large variations in view angles and human poses owing to the benefits of patch-wise graph matching. Extensive experiments on five benchmarks including VIPeR, Road, PRID450S, 3DPES and CUHK01 evidence the superior performance of GCT model over other state-of-the-art methods.
\end{abstract}

\section{Introduction}
\noindent Person re-identification (Re-ID), which aims to associate a probe image to each individual in a gallery set (usually across different non-overlapping camera views), plays a crucial role in various applications including video surveillance, human retrieval, etc. Despite great successes in recent years, accurate Re-ID remains challenging due to many factors such as large appearance changes in different camera views and heavy body occlusions. To deal with these issues, numerous Re-ID approaches are proposed ~\cite{FarenzenaBPMC10,KaranamLR15,salicency,sim_spatial_constraints,KISSME,SVMML,lomo,pcca,learn_to_rank,lfda,TMA}.

\begin{figure}[!t]
  \centering
  \includegraphics[width=0.8\linewidth]{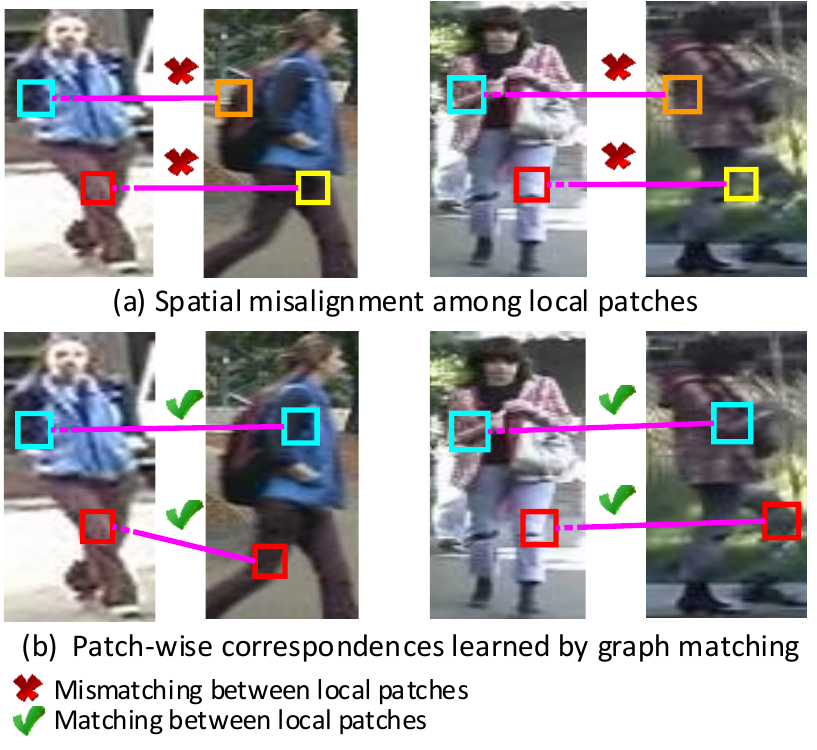}
  \caption{Illustration of misalignment problem in Re-ID. Image (a) shows misalignment among local patches caused by viewpoint changes. The proposed GCT model can capture the correct semantic matching among patches using patch-wise graph matching, as shown in image (b).}
  \label{GCT_insight}
\end{figure}

For Re-ID task, one major challenge is to deal with the inevitable spatial misalignments between image pairs caused by large variations in camera views and human poses, as shown in Fig. 1. Most existing methods~\cite{YangYYLYL14,kLFDA14,ChenYHZW15}, nevertheless, focus on addressing the problem of Re-ID by comparing the holistic differences between images, which ignores spatial misalignment problem. To alleviate this issue, there are some attempts to apply part-based approaches to handle misalignment~\cite{iccv15_correspondence,OreifejMS10,salicency,YangWLL17}. These methods divide objects into local patches and perform an {\it online} patch-level matching for Re-ID. Though these approaches can handle spatial misalignment to some extent, being in lack of spatial and visual context information among local patches, they still fail in presence of visually similar body appearances or occlusions.

\begin{figure*}[!t]
  \centering
  \includegraphics[width=0.82\linewidth]{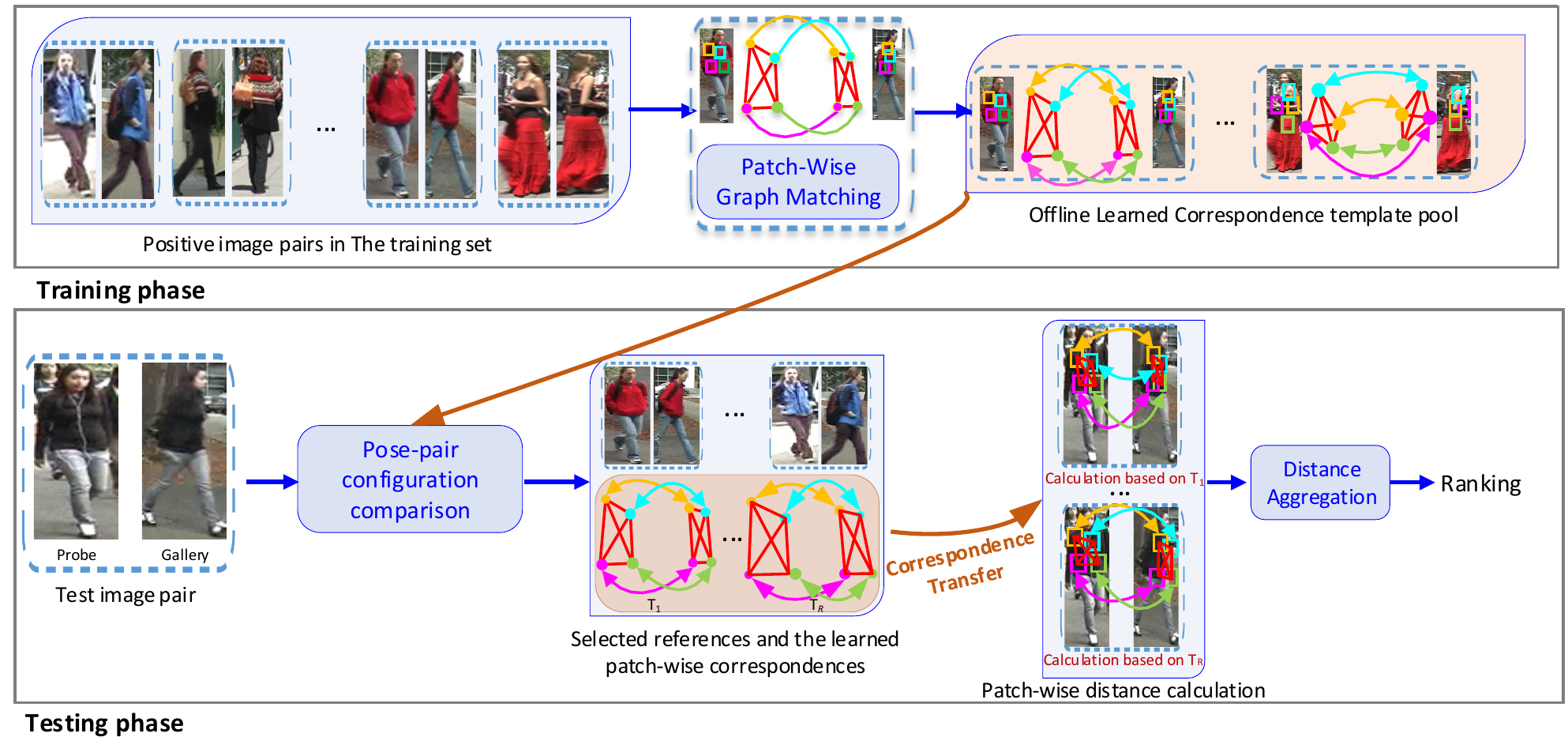}
  \caption{Illustration of GCT model. During training, spatial and visual context information are embedded into patch-wise graph matching to establish patch-level correspondences for different positive training pairs with various pose-pair configurations. During testing, for a pair of test samples, we use a simple pose-pair configuration comparison method to choose a few positive training pairs with the most similar pose-pair configurations as references, and then transfer the correspondences of these references to this test pair to compute local feature distances, which are then aggregated to calculate the overall feature distance.}\label{GCT_model}
\end{figure*}


In this paper, we propose to learn patch-level matching templates for positive training pairs via graph matching, and transfer the learned patch-level correspondences to test pairs with similar pose-pair configurations. In the formulation of graph matching, both spatial and visual context information are utilized to solve misalignment problem. The correspondence transfer in testing pahse is based on the observation that two image pairs with similar pose-pair configurations tend to share similar patch-level correspondences. Benefiting from part-based strategy and implicitly modeling body context information into graph matching procedure, our GCT algorithm is able to well deal with spatial misalignment. Besides, the regularization term imposed on the learned correspondences in patch-wise graph matching helps to improve the robustness of GCT model when occlusions occur. During testing, for each pair of samples, we present a simple but effective pose-pair configuration comparison method to select the training pairs with the most similar pose-pair configurations as references, and the learned patch-wise correspondences of these references are utilized to compute the overall feature distance between this test pair. Fig \ref{GCT_model} illustrates the details of GCT model. Experiments on five benchmarks show the effectiveness of GCT method.

In summary, we make the following contributions:
\begin{itemize}
\item Spatial and visual context information are utilized to exploit the patch-wise semantic correspondences between positive image pairs to handle spatial misalignment. And for the first time, a novel graph correspondence transfer (GCT) model is presented for person re-identification.
\item Based on the observation that image pairs with similar pose-pair configurations tend to share similar patch-level correspondences, we introduce a simple but effective pose-pair configuration comparison method to find the best references for each test image pair.
\item Extensive experiments on five challenging benchmarks demonstrate that the proposed GCT model performs favorably against state-of-the-art approaches, and in fact even better than many deep learning based solutions.
\end{itemize}

\section{Related Work}

Being extensively studied, numerous approaches have been proposed for Re-ID in recent years~\cite{ZhengSTWWT15}. Here we briefly review some related works of this paper.

To handle the problem of spatial misalignment, there are some attempts to apply part-based strategies for Re-ID. This kind of methods divide images into several parts and perform on-line patch-level matching to discard misalignments. In~\cite{lomo}, Liao {\it et al.} propose to incorporate body prior into Re-ID by decomposing human body into a few fixed stripes, however it still fails in the scenarios of large view changes. Cheng {\it et al.}~\cite{DEEP_MULTI_CHANNEL} take the advantages of body part detection, and propose a part-based appearance model to alleviate the influence of misalignment in Re-ID. However, this approach heavily relies on body part detection performance, leading to degradation in presence of occlusion. In~\cite{salicency}, Zhao {\it et al.} propose to mine the salient scores of different local patches for Re-ID by building dense correspondences between image pairs. The aforementioned algorithms, however, neglect body context information during feature designing or metric learning, resulting in performance deteriorating.

The most relevant work to ours is~\cite{iccv15_correspondence} (CSL), which aims to mine the patch-wise matching structure for each pair of cameras. Nevertheless, our GCT model differs from CSL in two aspects: (\romannumeral1) Instead of learning a holistic optimal correspondence structure for each camera pair in CSL, we utilize graph matching to establish accurate local feature correspondences for each positive image pair, and then transfer the learned correspondence templates for Re-ID. (\romannumeral2) We implicitly model the body context information, which is neglected in CSL, in the affinity matrix for graph matching to improve Re-ID performance in our work.

\section{The Proposed Approach}

In this section, we describe the detailed GCT model, which consists of correspondence learning with patch-wise graph matching in training phase, reference selection via pose-pair configuration comparison and patch-wise feature distance calculation and aggregation based on correspondence transfer.

\subsection{Patch-wise correspondence learning with graph matching}

To deal with the misalignment problem, a part-based strategy is adopted to represent human appearance. In specific, we decompose the images into many overlapping patches, and represent each image with an undirected attribute graph $G=(V, E, A^V)$, where each vertex $v_{i}$ in the vertex set $V=\{v_{i}\}_{i=1}^{n}$ denotes an image patch, and each edge encodes the context information of the connected vertex pair. $A^V = \{A^{V_P},A^{V_F}\}$ are vertex attributes representing spatial and visual features of the local patches.

During training, given a positive image pair $I_1$ and $I_2$ with identify labels $l_1$ and $l_2$, where $l_1=l_2$ (i.e., $I_1$ and $I_2$ belong to same person), they can be represented with attribute graphs $G_1=(V_1, E_1, A_1^V)$ and $G_2=(V_2, E_2, A_2^V)$, respectively. The patch-wise correspondence learning aims to establish the vertex correspondences $X \in \{0,1\}^{n_1\times n_2}$ between $V_1$ with $n_1$ vertexes and $V_2$ with $n_2$ vertexes, such that the intra-person similarity (i.e., $l_1 = l_2$) is maximized on the training set.

In Re-ID, $X_{ia}=1$ means the $i^{th}$ patch in $I_1$ semantically corresponds to the $a^{th}$ patch in $I_2$. Mathematically, the patch-wise correspondence learning is formulated as an Integer Quadratic Programming (IQP) problem as follows:
\begin{equation}
\begin{aligned}
& \arg \mathrm{\max\limits_\mathbf{x}} {\kern 1pt}{\kern 1pt}{\kern 1pt}{\kern 1pt}  {\kern 1pt}{\kern 1pt} \mathbf{x}^{\mathrm{T}}K\mathbf{x}, \\
s.t.& \begin{cases}
 {\kern 1pt}{\kern 1pt}{\kern 1pt} &X_{ia}\in\{0,1\} ,\forall i\in\{1,\cdots,n_1\} ,\forall a\in\{1,\cdots,n_2\} \\
&{\sum}_i X_{ia} \leq 1, \forall a\in\{1,\cdots,n_2\},\\
&{\sum}_a X_{ia} \leq 1, \forall i\in\{1,\cdots,n_1\},
\end{cases}
\end{aligned}
\label{eq:graph_matching_obj}
\end{equation}
where $\mathbf{x}=\mathrm{vec}(X)\in\{0,1\}^{n_1{\times}n_2}$ denotes the vectorized version of matrix $X$ and $K\in \mathbb{R} ^{{n_1n_2}\times{n_1n_2}}$ represents the corresponding affinity matrix between $G_1$ and $G_2$, which encodes the relational similarities between edges and vertices.

\subsubsection{Affinity matrix design}

Due to large variations in human body configuration caused by serious pose and view changes, it is not suitable to directly apply traditional spatial layout based affinity matrix for Re-ID. In addition, taking into consideration the importance of visual appearance in Re-ID, we combine both visual feature and spatial layout of human appearance to develop the affinity matrix.

In specific, the diagonal components $K^{ia,ia}$ of the affinity matrix $K$ (which capture the node compatibility between vertex $v_i \in V_1$ and vertex $v_a \in V_2$) are calculated as follows:

\begin{equation}
K^{ia,ia} = S_{ia}^P \cdot S_{ia}^F,
\end{equation}
where $S_{ia}^P$ and $S_{ia}^F$ refer to the {\it spatial proximity} and {\it visual similarity} between $v_i$ and $v_a$ respectively. The $S_{ia}^P$ and $S_{ia}^F$ can be mathematically computed with
\begin{equation}
\begin{aligned}
S_{ia}^P = \mathrm{exp}(-\| A_i^{V_P} - A_a^{V_P} \|_{2}), \\
S_{ia}^F = \mathrm{exp}(-\| A_i^{V_F} - A_a^{V_F} \|_{2}),
\end{aligned}
\end{equation}
where $A_i^{V_P}$ and $A_a^{V_P}$ denote spatial positions of $v_i$ and $v_a$, and $A_i^{V_F}$ and $A_a^{V_F}$ represent their visual features.

Likewise, for non-diagonal element $K^{ia,jb}$ in $K$, which encodes the compatibility between two edges $e_{ij}$ connecting $\{v_i \in V_1,v_j \in V_1\}$ and $e_{ab}$ connecting $\{v_a \in V_2,v_b \in V_2\}$, it can be obtained as the following
\begin{equation}
K^{ia,jb} = S_{ij,ab}^P \cdot S_{ij,ab}^F,
\end{equation}
where $S_{ij,ab}^P$ and $S_{ij,ab}^F$ represent spatial and visual compatibilities between edges $e_{ij}\in{E_1}$ and $e_{ab}\in{E_2}$, and they are calculated by
\begin{equation}
\begin{aligned}
S_{ij,ab}^P = \mathrm{exp}(-\| (A_i^{V_P} - A_j^{V_P}) - (A_a^{V_P} - A_b^{V_P}) \|_{2}), \\
S_{ij,ab}^F = \mathrm{exp}(-\| (A_i^{V_F} - A_j^{V_F}) - (A_a^{V_F} - A_b^{V_F}) \|_{2}).
\end{aligned}
\end{equation}
In this way, the calculated affinity matrix $K$ implicitly embeds the spatial and visual context information into the graph matching procedure, such that the matched vertices and edges have larger similarities and are more compatible with each other both spatially and visually. Therefore, we can obtain a satisfying patch-wise matching result for Re-ID.

\begin{figure}[!t]
  \centering
  \includegraphics[width=0.95\linewidth]{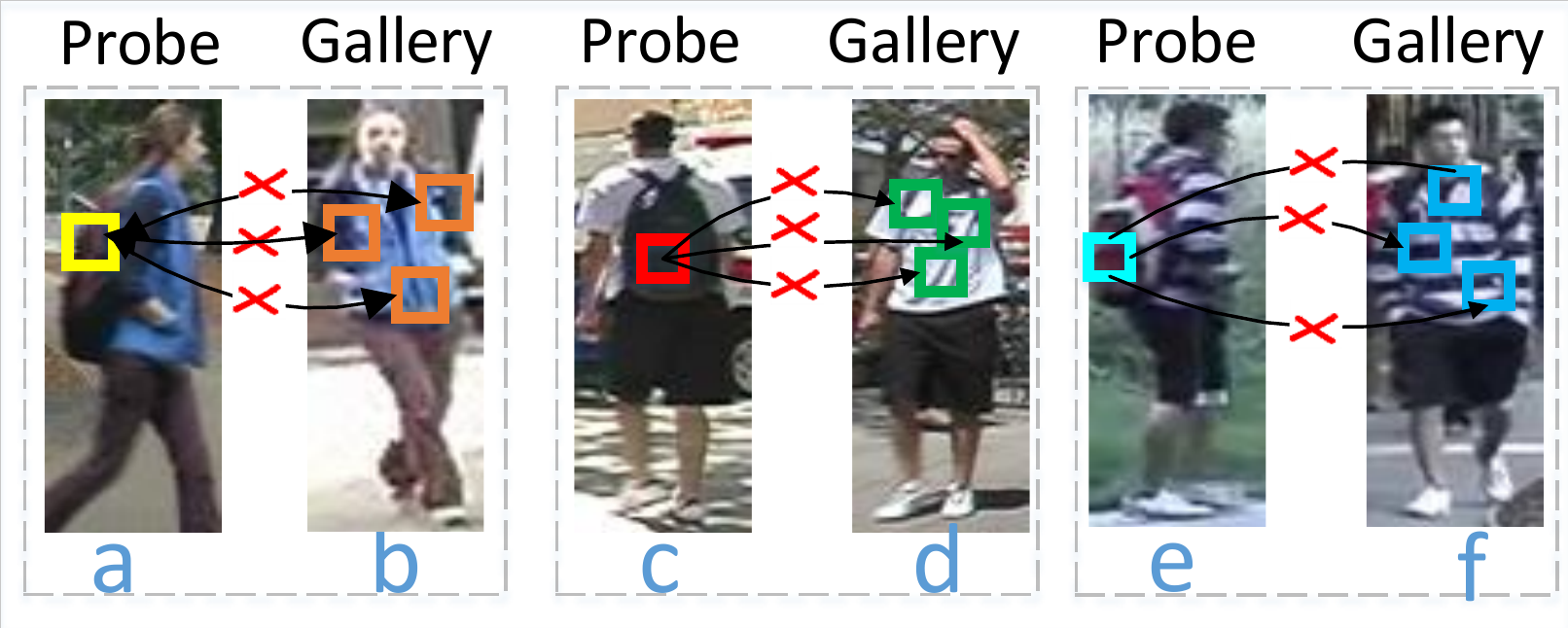}
  \caption{Sample images to demonstrate the fact that local patches visible in one view may not appear in the other.}
  \label{fig::invisible}
\end{figure}

\subsubsection{Outlier removal}

Due to spatial misalignment, some patches visible in one view may not appear in the other (see Figure~\ref{fig::invisible}). Therefore, imposing a global one-to-one match between patches could bring in noise and deteriorate the performance. In this case, only establishing correspondences between commonly visible parts of positive image pairs is more reasonable. To this end, we adopt the same strategy as in~\cite{SuhAL15} by incorporating a regularization term on the number of activated vertices. In this way, the probe patches that match with high spatial and visual similarities are activated, while the outliers that do not co-exist in the two images are excluded. Thus, the objective function (\ref{eq:graph_matching_obj}) can be rewritten as follows:
\begin{equation}
\begin{aligned}
& \arg  \mathrm{\max\limits_\mathbf{x}}  {\kern 1pt}{\kern 1pt}{\kern 1pt}{\kern 1pt}  {\kern 1pt}{\kern 1pt} \mathbf{x}^\mathrm{T}K\mathbf{x} - \lambda ||\mathbf{x}||_2^2, \\
s.t. &\begin{cases}
 {\kern 1pt}{\kern 1pt}{\kern 1pt} &X_{ia}\in\{0,1\} ,\forall i\in\{1,\cdots,n_1\} ,\forall a\in\{1,\cdots,n_2\} \\
&{\sum}_i X_{ia} \leq 1, \forall a\in\{1,\cdots,n_2\},\\
&{\sum}_a X_{ia} \leq 1, \forall i\in\{1,\cdots,n_1\},
\end{cases}
\end{aligned}
\label{eq:graph_matching_obj_final}
\end{equation}
where $\lambda$ is a trade-off parameter to control the difficulty of a new probe vertex being activated. Larger $\lambda$ means more extra similarity should be brought to activate a new vertex. We adopt the method in~\cite{SuhAL15} to solve Eq.(\ref{eq:graph_matching_obj_final}).

In existing part-based Re-ID methods ~\cite{iccv15_correspondence,salicency}, an image is typically decomposed into hundreds of patches to capture detailed local visual information, leading to intractability in solving Eq. (\ref{eq:graph_matching_obj_final}). To reduce the search space and inhibit potential matching ambiguity, we adopt the commonly utilized spatial constraints~\cite{lomo,sim_spatial_constraints} to lower the computational load, as well as to improve the patch-wise matching results. More specifically, a probe image is divided into a few horizontal stripes, and for each stripe in the probe image, patch-wise correspondences are established between the corresponding gallery stripe within the search range in the gallery image by optimizing Eq.(\ref{eq:graph_matching_obj_final}).

By optimizing Eq.(\ref{eq:graph_matching_obj_final}), we can obtain a set of graph correspondence templates for the positive image pairs in the training set.

\subsection{Reference selection via pose-pair configuration comparison}
\label{ref_selection}
We argue that the learned patch-wise correspondence patterns can be favorably transferred to image pairs with similar pose-pair configurations in the testing set, and these transferred correspondences are directly utilized to compute the distance between probe and gallery images. To this end, we need to find out the best references for each pair of test images from the training set. Since pose configurations are closely related to body orientations, we calculate the similarities between different pose pairs by comparing their related body orientations.
\begin{figure}[!t]
  \centering
  \includegraphics[width=0.95\linewidth]{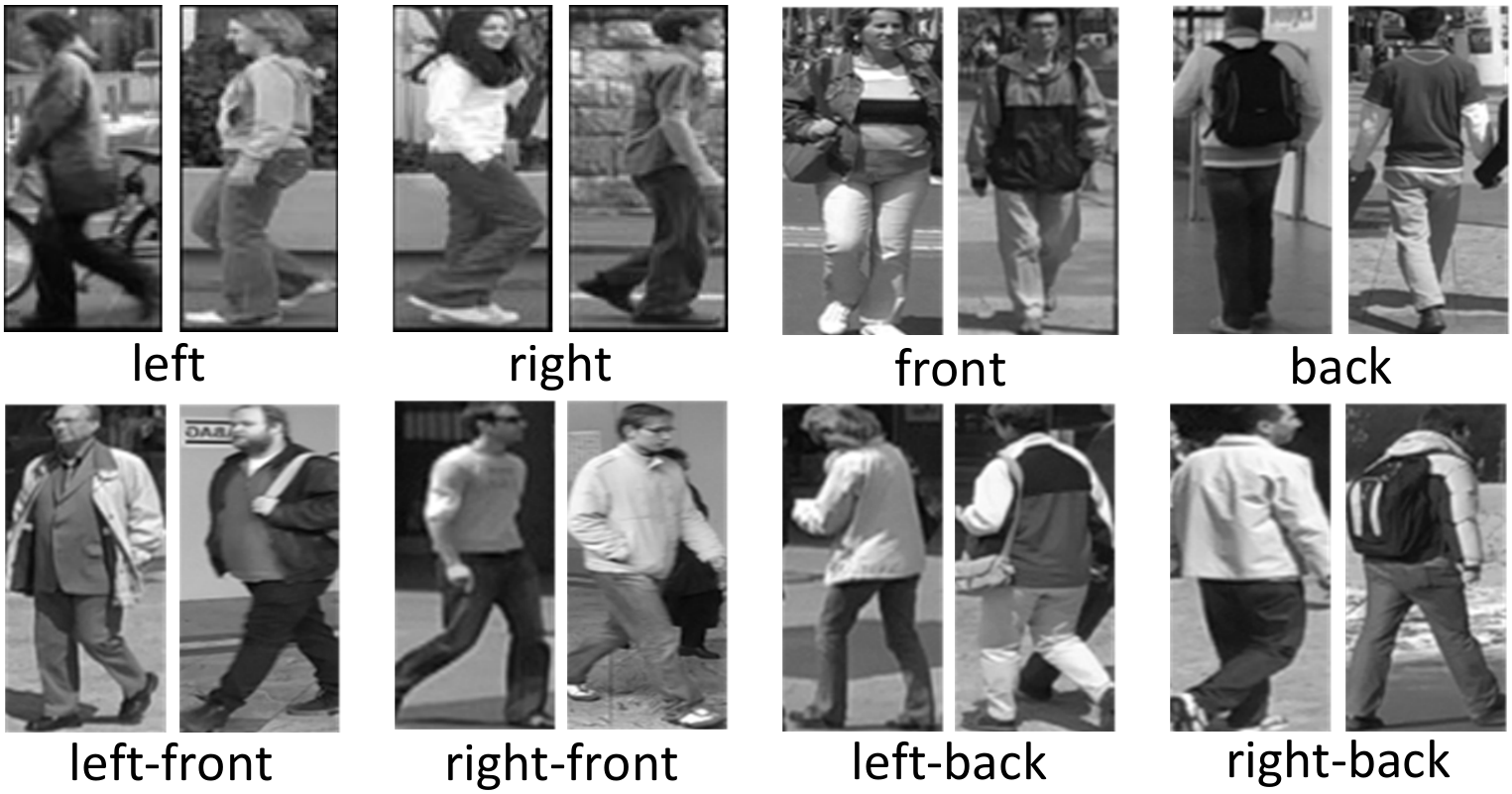}
  \caption{Sample images in eight classes in the TUD dataset~\cite{TUD_DATASET}. Note that the TUD dataset is only used for training the reference selection model, and is different from the benchmarks in experiments.}
  \label{fig:multi_view_img}
\end{figure}

We propose to utilize a simple but effective random forest method~\cite{random_forest} to compare different body orientations. Specifically, images are classified into eight different clusters including `left', `right', `front', `back', `left-front', `right-front', `left-back' and `right-back', according to their body orientations, as shown in Figure~\ref{fig:multi_view_img}. In order to train the random forest model, each image is represented with multi-level HoG features (i.e., cell sizes are set to $8 \times 8, 16\times 16, 32\times 32$ respectively, with a block size of $2 \times 2$ cells and a block stride of one cell for each direction), and then fed into each decision tree to build the random forest~\cite{random_forest}. Once the random forest $M = \{tree_1,tree_2,\cdots,tree_T\}$ is built, where $T$ denotes the number of trees in $M$, the body orientation proximity $\mathrm{O}$ between two images $I_i$ and $I_j$ can be calculated as:
\begin{equation}
\mathrm{O}(I_i,I_j) = \frac {1}{T}\sum\limits_{t = 1}^T{y_{ij}^t},
\label{eq:orientaiton_proximity}
\end{equation}
where $y_{ij}^t$ is an indicator, and $y_{ij}^t = 1$ if $I_i$ and $I_j$ arrive at the same terminal node in $tree_t \in M$, otherwise $y_{ij}^t = 0$.

Given two image pairs $P=(I_p, I_g)$ and $P'=(I_p^{'}, I_g^{'})$, their pose-pair configuration similarity $S(P, P')$ is computed as follows
\begin{equation}
S(P, P') = \mathrm{O}(I_p,I_p^{'}) \cdot \mathrm{O}(I_g,I_g^{'})
\label{eq:body_sim}
\end{equation}

\begin{figure}[!t]
  \centering
  \includegraphics[width=0.95\linewidth]{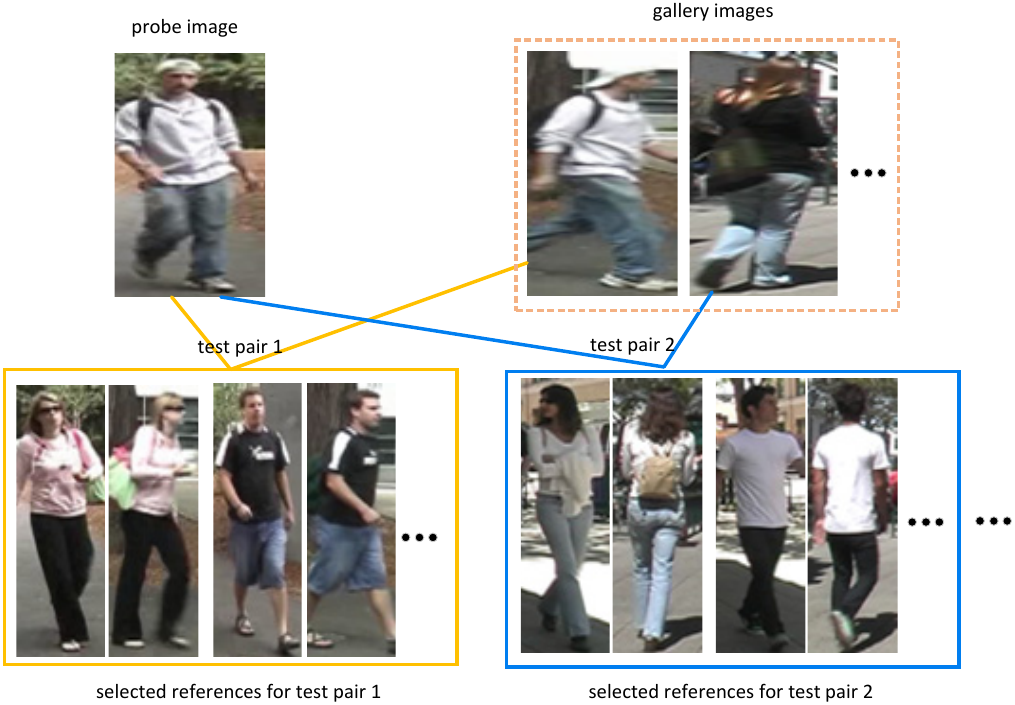}
  \caption{Demonstration of reference selection results.}
\label{fig:proximity_results}
\end{figure}

With Eq.(\ref{eq:body_sim}), we can calculate the body configuration similarities between an test image pair and all the positive training image pairs, and select $R$ training image pairs with highest similarities as the best references for the test pair. Figure~\ref{fig:proximity_results} shows some selected references of the sample test pairs.

\subsection{Distance calculation and aggregation with correspondence transfer}

Given that image pairs with similar pose-pair configurations tend to share similar patch-level correspondences, for each test pair of images, we propose to transfer the matching results of the selected references (the way to select the references is presented in Section~\ref{ref_selection}) to calculate the patch-wise feature distances of this test pair. The details of feature distance calculation using the selected references are presented in the following part.

After obtaining the best references for each test image pair, we transfer their learned correspondences to compute the distance between the test images. Given a pair of test images $\bar{P}=(\bar{I}_p, \bar{I}_g)$, where $\bar{I}_p$ and $\bar{I}_g$ are the probe and gallery image respectively, we can choose $R$ references for $\bar{P}$ as described in Section 3.2. Let $\mathcal{T}=\{ T_i \}_{i=1}^{R}$ represent the correspondence template set of these $R$ references, where each template $T_{i}=\{c_{ij}\}_{j=1}^{Q_{i}}$ contains $Q_{i}$ patch-wise correspondences, and each correspondence $c_{ij}=(w_{ij}^{p}, w_{ij}^{g})$ denotes the positions of matched local patches in the probe and gallery image.

For the test pair $\bar{P}$, we can compute the distance $D$ between $\bar{I}_p$ and $\bar{I}_g$ as the following:
\begin{equation}
D(\bar{I}_p, \bar{I}_g) = \sum\limits_{i=1}^{R}\sum\limits_{j=1}^{Q_{i}}{\delta(f_{p}^{w_{ij}^{p}},f_{g}^{w_{ij}^{g}})}
\label{rank}
\end{equation}
where $\delta(\cdot,\cdot)$ denotes the KISSME metric~\cite{KISSME}, and $f_{p}^{w_{ij}^{p}}$ and $f_{g}^{w_{ij}^{g}}$ represent features of local patches located at $w_{ij}^{p}$ and $w_{ij}^{g}$ in probe image $\bar{I}_p$ and gallery image $\bar{I}_g$. In this paper, we use Local Maximal Occurrence features~\cite{lomo} to represent each image patch.

With Eq.(\ref{rank}), we can calculate the patch-wise feature distances between each correspondence (a semantically matched patch pair between the probe and gallery images) of the selected references, these local feature distances are then equally aggregated to obtain the overall distance between the test image pairs. The gallery image with the smallest distance is determined to be the re-identifying result.

\section{Experimental Results}
\subsection{Experimental setup}

\textbf{Datasets} We conduct extensive experiments on three challenging single-shot datasets (VIPeR, Road and PRID450S), and two multi-shot datasets (3DPES and CUHK01). The characteristics of each dataset are detailed as follows:

\textbf{VIPeR dataset:} The VIPeR~\cite{viper} dataset consists of 632 people with two images from two cameras for each person. It bears great variations in poses and illuminations, most of the image pairs contain viewpoint changes larger than 90 degrees.

\textbf{Road dataset:} The Road dataset~\cite{iccv15_correspondence}, consisting of 416 image pairs, is captured from a realistic crowd road scene, with serious interferences from occlusions and large pose variations, making it quite challenging.

\textbf{PRID450S dataset:} The PRID 450S~\cite{prid_450s} dataset contains 450 pairs of images from two camera views. The very similar appearances in images make it very challenging for person re-identification.

\textbf{3DPES dataset:} The 3DPES dataset~\cite{three_dpes} contains 1011 images of 192 persons captured from 8 different cameras. The number of images for a specific person ranges from 2 to 26, and the bounding boxes are generated from automatic pedestrian detection.

\textbf{CUHK01 dataset:} The CUHK01 dataset~\cite{CUHK01} is a medium-sized dataset for Re-id, captured from two disjoint camera views. It consists of 971 individuals, with each person having two images under each camera view. Different from VIPeR, images in CUHK01 are of higher resolutions. On this dataset, we adopt the commonly utilized 485/486 setting for performance evaluation.

\noindent
\textbf{Parameter setup} The proposed algorithm is implemented in Matlab on an Intel(R) Core(TM) i7-5820K CPU of 3.30GHz. The $\lambda$ in Eq.(\ref{eq:graph_matching_obj_final}) is set to 2. The number of trees in the random forest model is 500. The best $R$ for VIPeR, Road, PRID450S, 3DPES and CUHK01 datasets are 20, 5, 10, 20 and 20 respectively. The size of local patch is $32 \times 24$. All the parameters will be available in the source code to be released for accessible reproducible research.

\begin{figure}[!t]
  \centering
  \includegraphics[width=\linewidth]{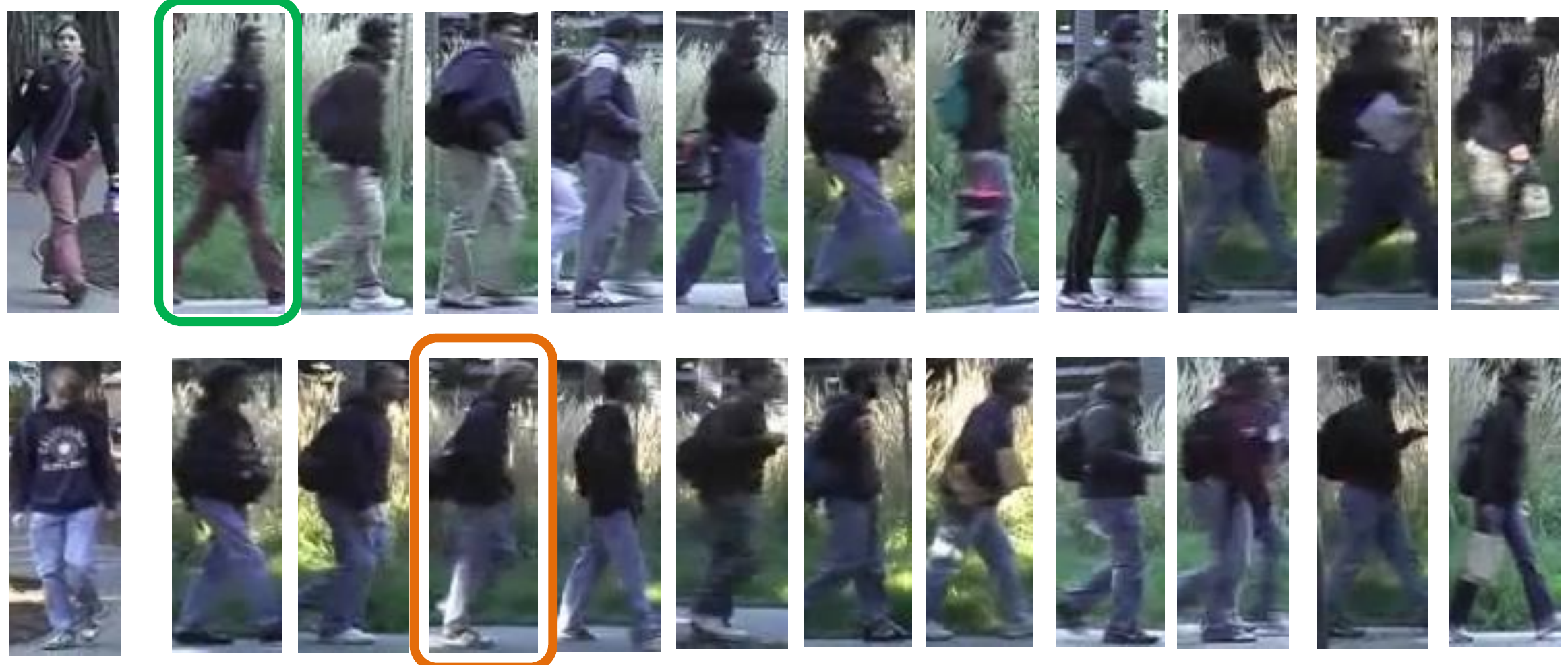}
  \caption{Top ranked images selected by our GCT algorithm. The first column are two probe images, and the following are the top ranked gallery images of each probe obtained by GCT. Images marked by green/orange bounding boxes in each row are the ground-truth matches of each probe.}
\label{fig:viper_results}
\end{figure}

\noindent
\textbf{Evaluation} We adopt the common half-training and half-testing setting~\cite{KISSME}, and randomly split the dataset into two equal subsets. The training/testing sets are further divided into the probe and gallery sets according to their view information. On all the datasets, both the training/testing set partition and probe/gallery set partition are performed 10 times and average performance is recorded. The performance is evaluated by cumulative matching characteristic (CMC) curve, which represents the expected probability of finding the correct match for a probe image in the top $r$ matches in the gallery list.

We record the top ranked gallery images of some sample probe images on the VIPeR dataset, which is presented in Figure~\ref{fig:viper_results}. As shown in Figure~\ref{fig:viper_results}, the proposed GCT algorithm can successfully rank images visually similar to the probe image ahead of others, which is the key requirement of most existing surveillance systems. Please note the second row of Figure~\ref{fig:viper_results}, the top ranked gallery images by GCT are all with dark coats and blue jeans, and to some extent, the rank1 image is visually more similar than the correct match marked by orange bounding box w.r.t. the probe image. Therefore, the proposed GCT algorithm is able to handle the spatial misalignment problem and generate satisfying ranking results for practical applications.

\subsection{Overall performance}

The GCT algorithm is implemented in Matlab on a PC with i7-5820K. The average training and testing time for an image pair are 0.096s and 0.0024s. To validate our approach, we evaluate the GCT model on five challenging datasets and compare it with several state-of-the-art approaches. The detailed comparison results are presented as follows.

\begin{table}[!t]\small
	\caption{Comparisons of top $r$ matching rate using CMC (\%) on VIPeR dataset. The best and second best results are marked in red and blue, respectively.
		\label{tab:VIPeRDataset}}
	\begin{center}
		\begin{tabular}{@{}C{3cm}@{}| @{}C{1.2cm}@{}@{}C{1.2cm}@{}@{}C{1.2cm}@{}@{}C{1.2cm}@{}}
			\hline
			Methods  & r=1 & r=5 & r=10 & r=20 \\
			\hline
			\hline
			SalMatch &  $30.2$ & $52.3$ & $65.5$ & $79.2$\\
			Semantic &  $41.6$ & $71.9$ & $86.2$ & $\textcolor{blue}{\bf 95.1}$\\
			LSSCDL &  $42.7$ & $-$& $84.3$  & $91.2$\\
			KISSME & $27.3$ & $55.3$ & $69.0$ & $82.7$\\
			SVMML & $30.0$ & $64.7$ & $79.0$ & $91.3$\\
			kLFDA &  $32.4$ & $65.9$ & $79.8$ & $90.8$\\
			Polymap & $36.8$ &$70.4$ &$83.7$ & $91.7$ \\
			LMF+LADF & $43.4$ & $73.0$ & $84.9$ & $93.7$ \\
			LOMO+XQDA &  $40.0$ & $68.1$ & $80.5$ & $91.1$\\
			DCSL &  $44.6$ & $73.4$ & $82.6$ & $-$\\
			TMA&  $48.2$ & $-$ & $\textcolor{red}{\bf87.7}$ & $\textcolor{red}{\bf 95.5}$\\
			TCP &  $47.8$ & $\textcolor{blue}{\bf 74.7}$ & $84.8$ & $91.1$\\
			DGD &  $35.4$ & $62.3$ & $69.3$ & $-$\\
			Spindle-Net &  $\textcolor{red}{\bf 53.8}$ & $74.1$ & $83.2$ & $92.1$\\
			CSL &  $34.8$ & $68.7$ & $82.3$ & $91.8$\\
			\hline
			\hline
			Our GCT&  $\textcolor{blue}{\bf 49.4}$ & $\textcolor{red}{\bf 77.6}$ & $\textcolor{blue}{\bf 87.2}$ & $94.0$\\\hline
		\end{tabular}
	\end{center}
\end{table}

\begin{table}[!t]\small
	\caption{Comparison of top $r$ matching rate using CMC (\%) on Road dataset. The best and second best results are marked in red and blue, respectively.
		\label{Road_Dataset.}}
	\begin{center}
		\begin{tabular}{@{}C{3cm}@{}| @{}C{1.2cm}@{}@{}C{1.2cm}@{}@{}C{1.2cm}@{}@{}C{1.2cm}@{}}
			\hline
			Methods  & r=1 & r=5 & r=10 & r=20 \\
			\hline\hline
			eSDC-knn &  $52.4$ & $74.5$ & $83.7$ & $89.9$\\
			CSL &  $\textcolor{blue}{\bf 61.5}$ & $\textcolor{blue}{\bf91.8}$ & $\textcolor{blue}{\bf 95.2}$ & $\textcolor{blue}{\bf98.6}$\\
			\hline\hline
			Our GCT &  $\textcolor{red}{\bf 88.8}$ & $\textcolor{red}{\bf 96.7}$ & $\textcolor{red}{\bf 98.4}$ & $\textcolor{red}{\bf 99.6}$\\
			\hline
		\end{tabular}
	\end{center}
\end{table}

On the VIPeR dataset, we compare the GCT with other fifteen algorithms, including SalMatch~\cite{salicency}, Semantic~\cite{Semantic15}, LSSCDL~\cite{LSSCDL16}, KISSME~\cite{KISSME}, SVMML~\cite{SVMML}, kLFDA~\cite{kLFDA14}, Polymap~\cite{ChenYHZW15}, LMF+LADF~\cite{mid_filter14}, LOMO+XQDA~\cite{lomo}, DCSL~\cite{DCSL}, TMA~\cite{TMA}, TCP~\cite{DEEP_MULTI_CHANNEL}, DGD~\cite{DGD}, Spindle Net~\cite{Spindle_Net} and CSL~\cite{iccv15_correspondence}. The comparison results are presented in Table~\ref{tab:VIPeRDataset}. As illustrated in Table~\ref{tab:VIPeRDataset}, the proposed GCT algorithm achieves the best recognition rate at rank 5, and competitive performances at rank 1, 10 and 20.

The Road dataset is proposed in CSL~\cite{iccv15_correspondence}. For comprehensive comparison, we also report the result on this dataset, and compare it with eSDC-knn~\cite{salicency} and CSL~\cite{iccv15_correspondence}. As shown in Table~\ref{Road_Dataset.}, compared to CSL~\cite{iccv15_correspondence}, our algorithm obtains significant improvements of 27.3 percents on rank 1 recognition rate, 4.9 percents on rank 5, 3.2 percents on rank 10 and 1 percent on rank 20 recognition rate, respectively.

\begin{table}[!t]\small
	\caption{Comparison of top $r$ matching rate using CMC (\%) on PRID450S dataset. The best and second best results are marked in red and blue, respectively.
		\label{tab:prid_Dataset.}}
	\begin{center}
		\begin{tabular}{@{}C{3cm}@{}| @{}C{1.2cm}@{}@{}C{1.2cm}@{}@{}C{1.2cm}@{}@{}C{1.2cm}@{}}
			\hline
			Methods  & r=1 & r=5 & r=10 & r=20 \\
			\hline\hline
			KISSME &  $33$ & $-$ & $71$ & $79$\\
			SCNCDFinal& $41.6$ & $68.9$ & $79.4$ & $87.8$\\
			Semantic &  $44.9$ & $71.7$ & $77.5$ & $86.7$\\
			TMA & $\textcolor{blue}{\bf54.2}$ & $\textcolor{blue}{\bf73.8}$ & $\textcolor{blue}{\bf83.1}$ & $\textcolor{red}{\bf90.2}$\\
			NSFT & $40.9$ & $64.7$ & $73.2$ & $81.0$\\
			CSL &  $44.4$ & $71.6$ & $82.2$ & $\textcolor{blue}{\bf 89.8}$\\
			\hline\hline
			Our GCT &  $\textcolor{red}{\bf 58.4}$ & \textcolor{red}{\bf 77.6} & \textcolor{red}{\bf 84.3} & \textcolor{blue}{\bf 89.8}\\
			\hline
		\end{tabular}
	\end{center}
\end{table}

\begin{table}[!t]
	\caption{Comparison of top $r$ matching rate using CMC (\%) on 3DPES dataset. The best and second best results are marked in red and blue, respectively.
		\label{tab:3dpes_Dataset.}}
	\begin{center}
		\begin{tabular}{@{}C{3cm}@{}| @{}C{1.2cm}@{}@{}C{1.2cm}@{}@{}C{1.2cm}@{}@{}C{1.2cm}@{}}
			\hline
			Methods  & r=1 & r=5 & r=10 & r=20 \\
			\hline\hline
			LFDA &  $45.5$ & $69.2$ & $-$ & $86.1$\\
			ME & $53.3$ & $76.8$ & $-$ & $92.8$\\
			kLFDA &  $54.0$ & $77.7$ & $85.9$ & $92.4$\\
			PCCA &  $41.6$ & $70.5$ & $81.3$ & $90.4$\\
			rPCCA &  $47.3$ & $75.0$ & $84.5$ & $91.9$\\
			SCSP &  $57.3$ & $79.0$ & $-$ & $91.5$\\
			WARCA &  $51.9$ & $75.6$ & $-$ & $-$\\
			DGD &  $56.0$ & $-$ & $-$ & $-$\\
			Spindle-Net &  $\textcolor{blue}{\bf62.1}$ & $\textcolor{blue}{\bf83.4}$ & $\textcolor{blue}{\bf90.5}$ & $\textcolor{blue}{\bf95.7}$\\
			CSL &  $57.9$ & $81.1$ & $89.5$ & $93.7$\\
			\hline\hline
			Our GCT &  $\textcolor{red}{\bf 69.8}$ & $\textcolor{red}{\bf 92.4}$ & $\textcolor{red}{\bf 95.5}$ & $\textcolor{red}{\bf 97.2}$\\
			\hline
		\end{tabular}
	\end{center}
\end{table}

On the PRID450S dataset, we compare with six state-of-the-arts algorithms, including KISSME~\cite{KISSME}, SCNCDFinal~\cite{YangYYLYL14}, Semantic~\cite{Semantic15}, TMA~\cite{TMA}, NSFT~\cite{NSFT} and CSL~\cite{iccv15_correspondence}. As shown in Table~\ref{tab:prid_Dataset.}, our algorithm achieves the best results on recognition rates of small ranks, which is more critical in practical use.

\begin{figure*}[!htb]
\centering
\begin{tabular}{@{}C{3.6cm}@{}C{3.6cm}@{}C{3.6cm}@{}C{3.6cm}@{}C{3.6cm}@{}}
\includegraphics[width= 3.58cm]{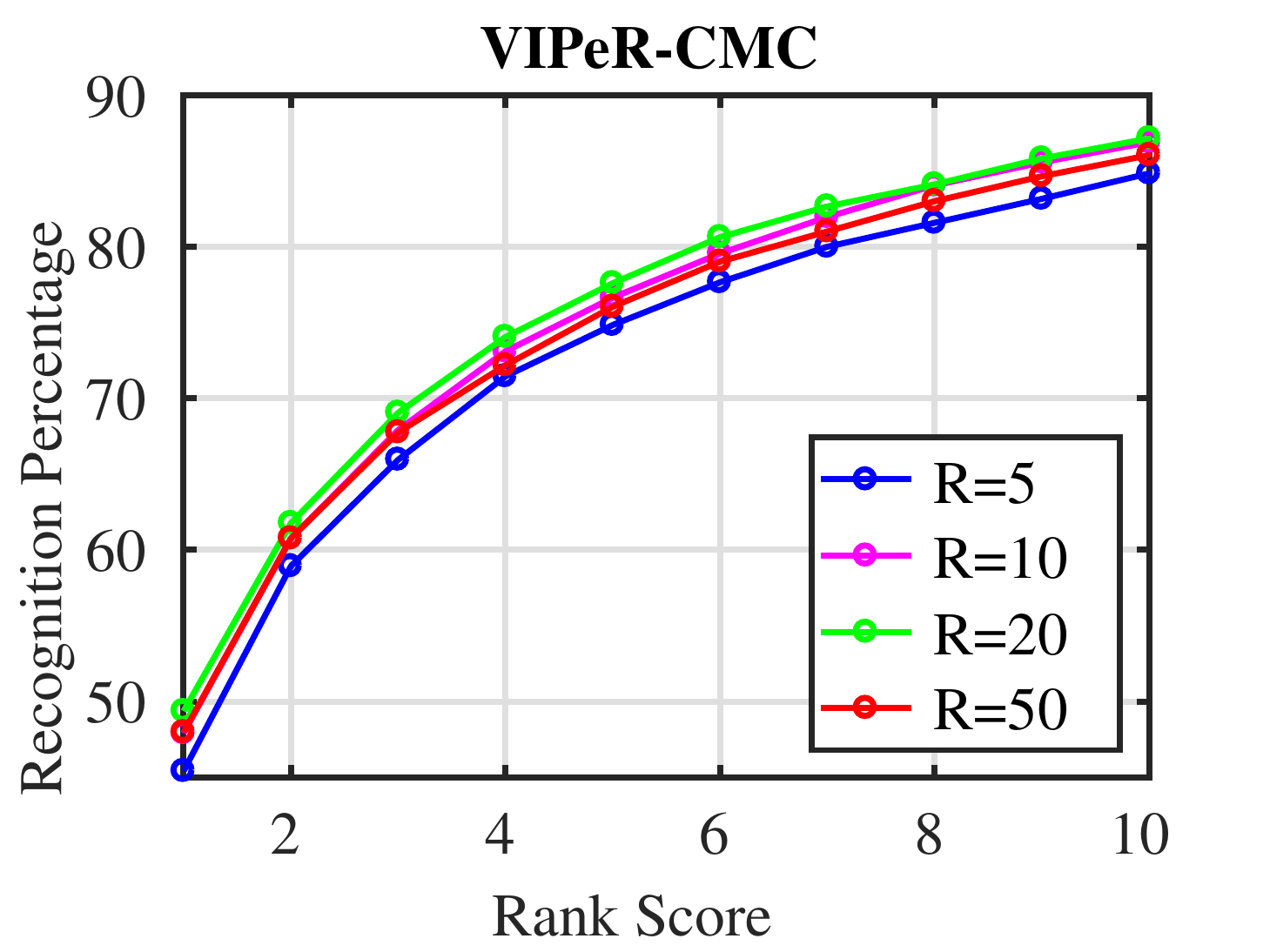}&\includegraphics[width=3.58cm]{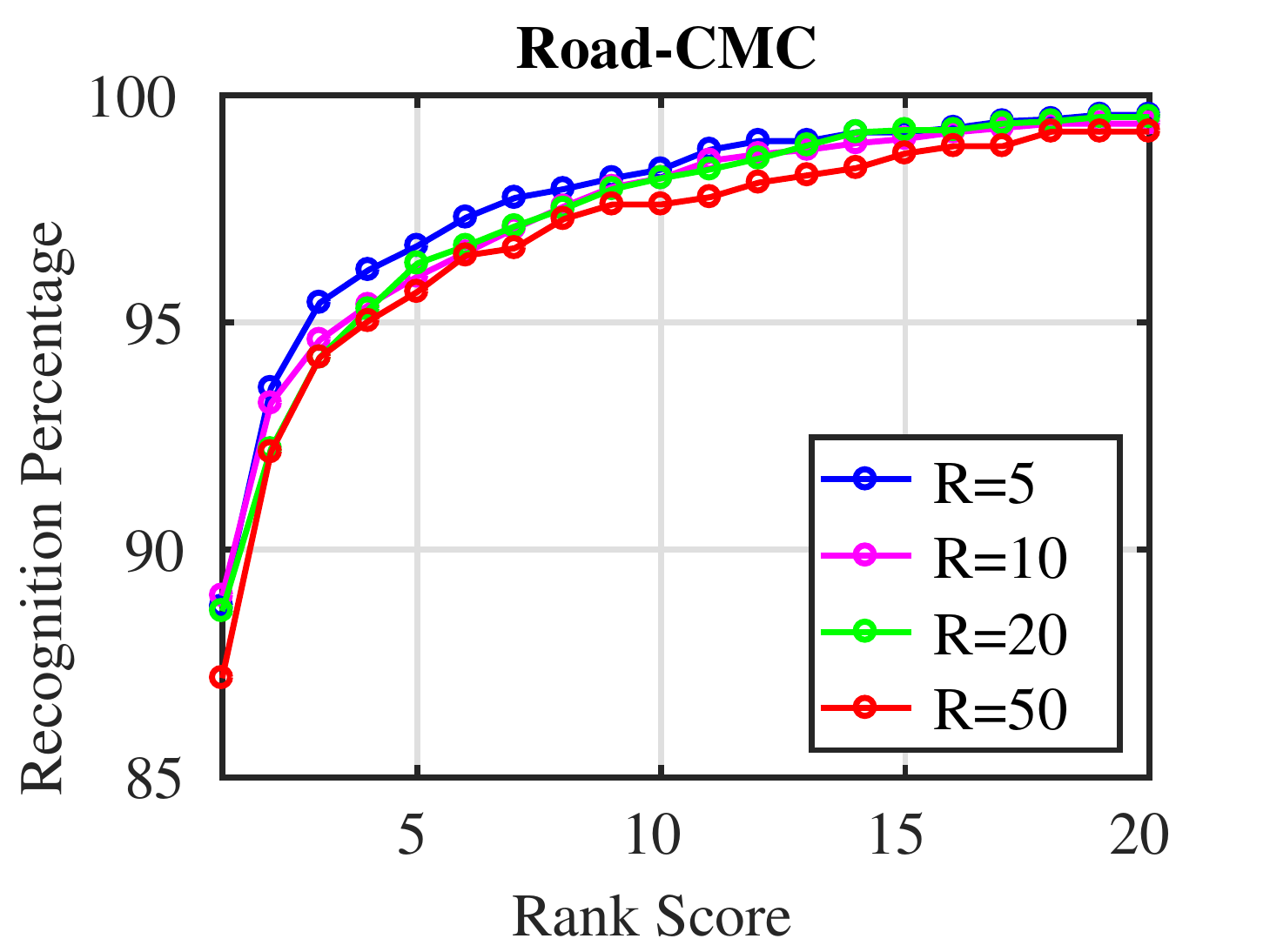}&
\includegraphics[width=3.58cm]{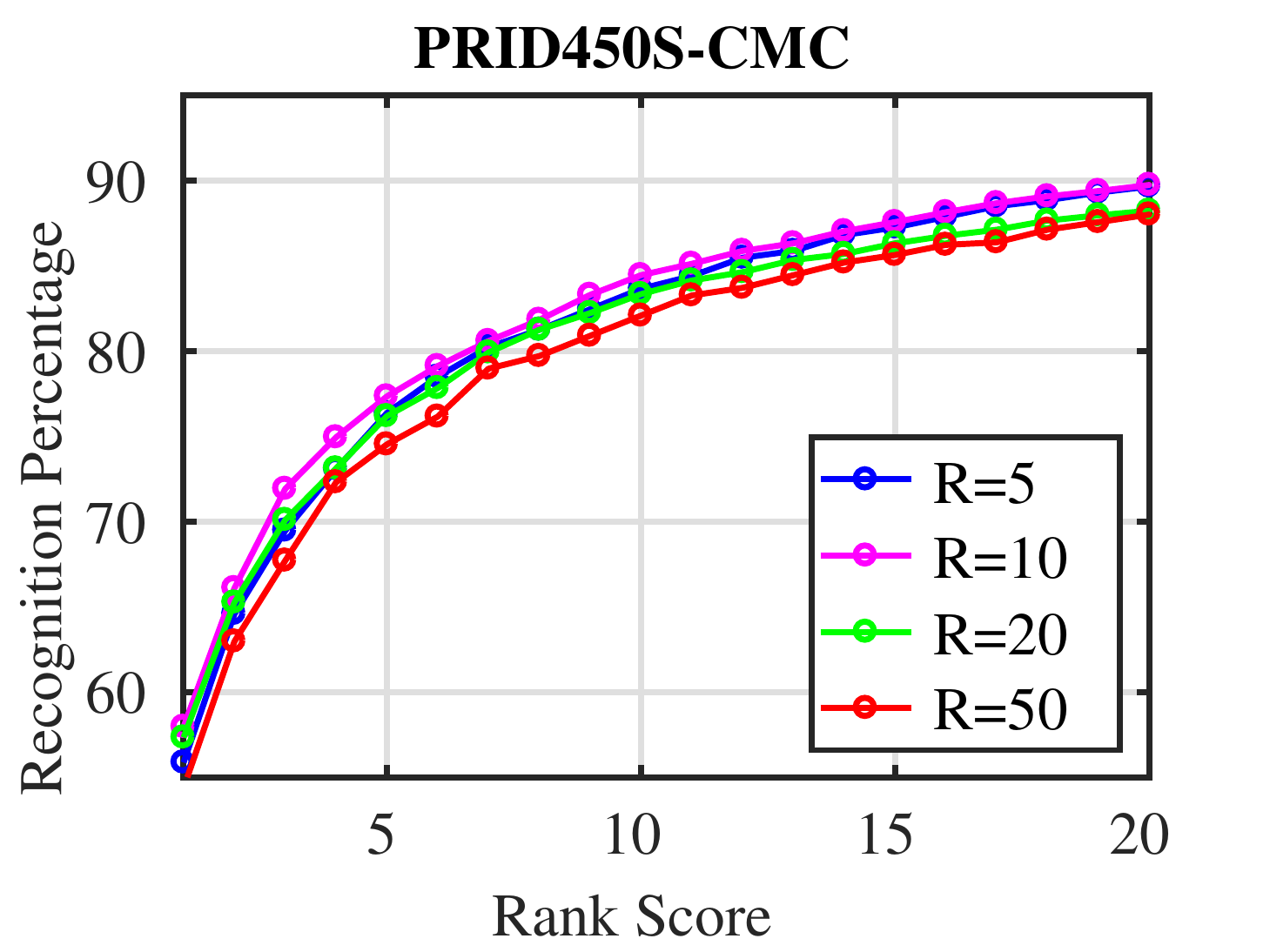}&\includegraphics[width=3.58cm]{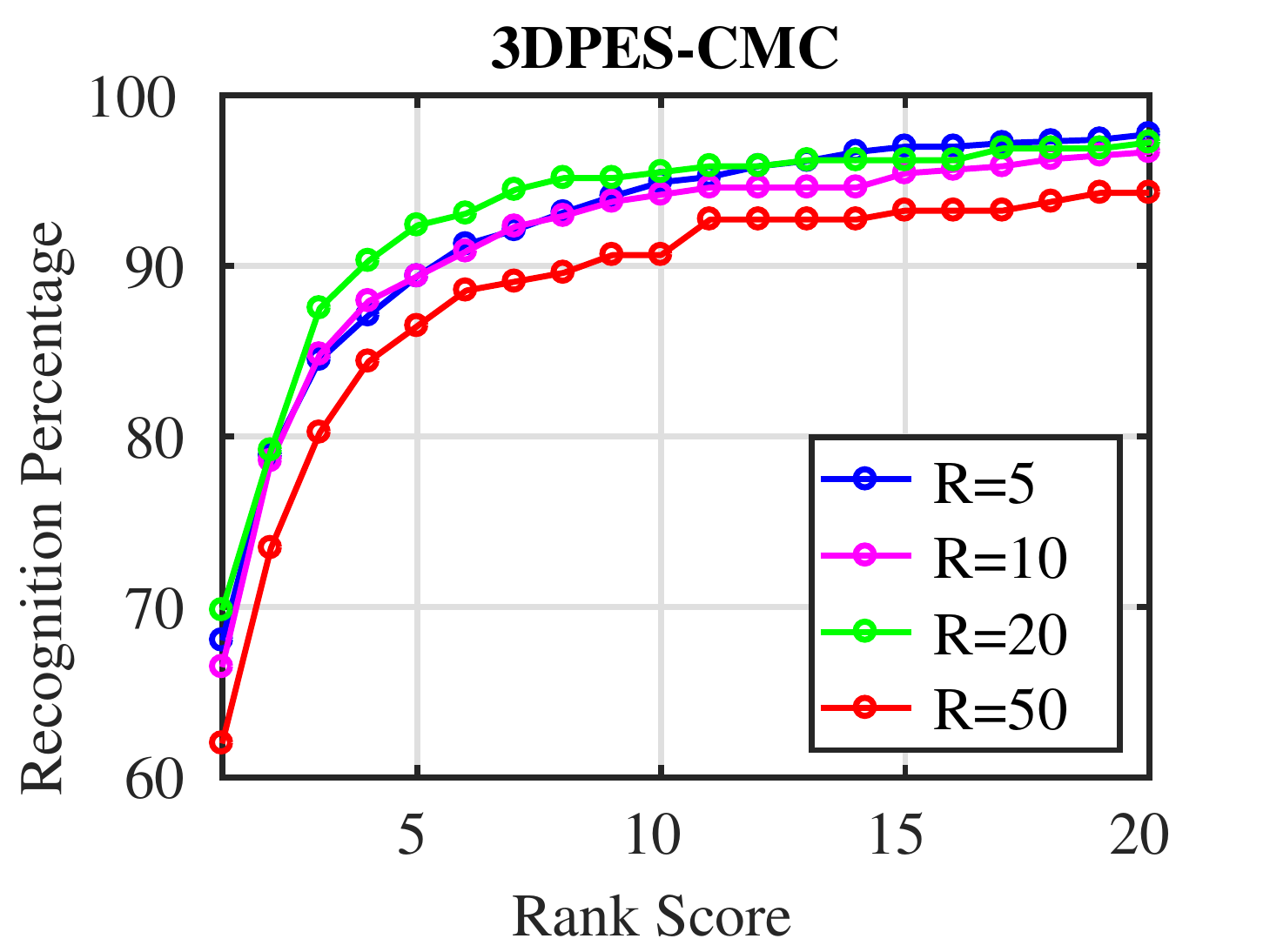}&\includegraphics[width=3.58cm]{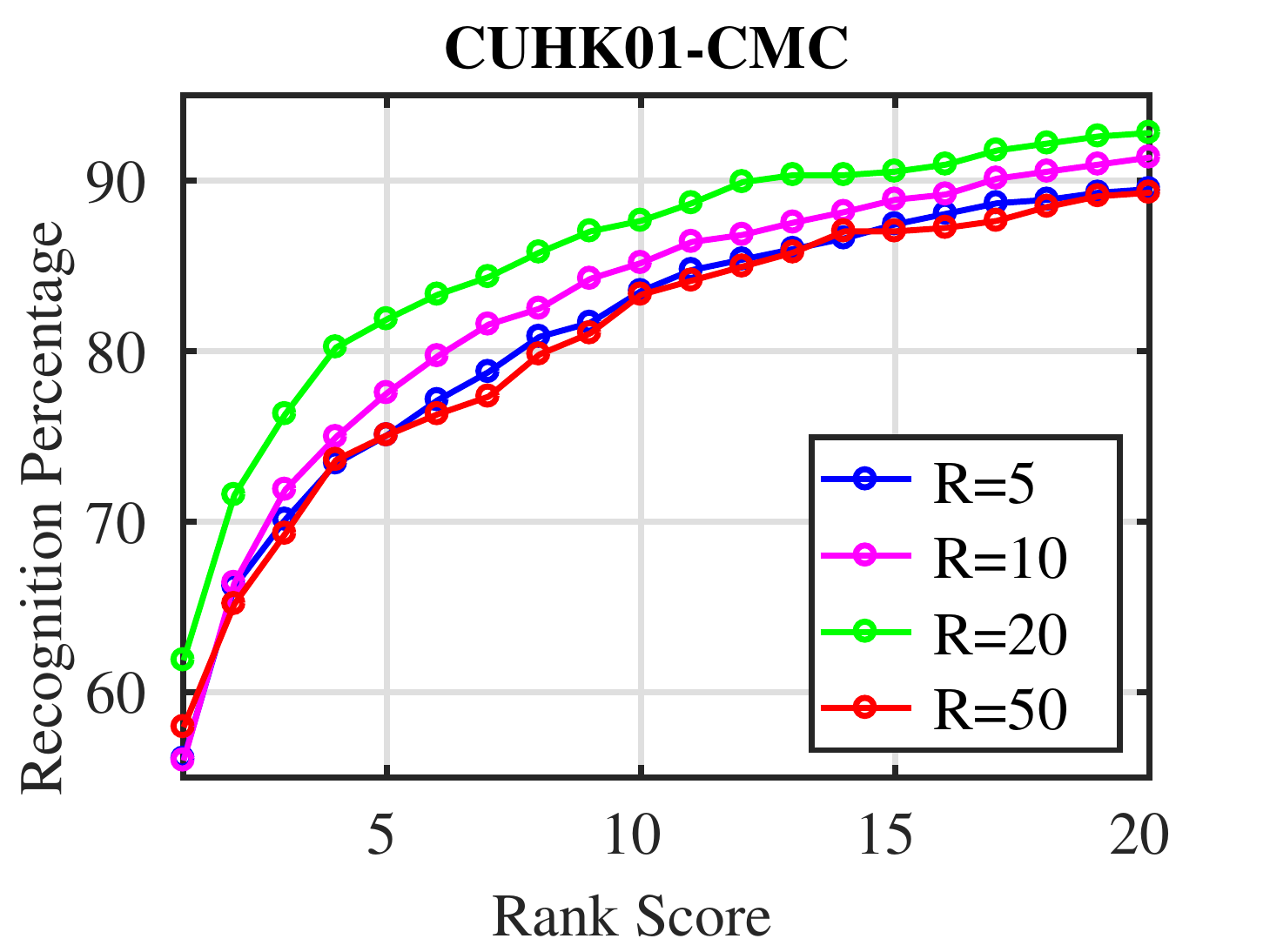}\\
\small{(a) VIPeR} & \small{(b) Road} &\small{(c) PRID450S} & \small{(d) 3DPES}  & \small{(e) CUHK01} \\
\end{tabular}
\caption{Evaluation of different numbers of selected references $R$ on the five datasets.}\label{fig:knn results}
\end{figure*}

On the 3DPES dataset, we compare the GCT method with ten algorithms, including LFDA~\cite{lfda}, ME~\cite{learn_to_rank}, kLFDA~\cite{kLFDA14}, PCCA~\cite{pcca}, rPCCA~\cite{kLFDA14}, SCSP~\cite{sim_spatial_constraints}, WARCA~\cite{WARCA}, DGD~\cite{DGD}, Spindle Net~\cite{Spindle_Net} and CSL~\cite{iccv15_correspondence}. As shown in Table~\ref{tab:3dpes_Dataset.}, we can see that the proposed algorithm outperforms existing state-of-the-art algorithms, and even deep learning based algorithm~\cite{Spindle_Net}. Note that the images in this dataset are automatic detection results from videos captured under eight cameras, bringing serious pose variations, illumination changes and scale variations. With the help of the patch-wise graph matching, our GCT model is robust to deal with these issues.

On the CUHK01 dataset, we compare with nine state-of-the-art algorithms, including Semantic~\cite{Semantic15},  kLFDA~\cite{kLFDA14}, IDLA\cite{AhmedJM15}, DeepRanking~\cite{deep_ranking}, ME~\cite{learn_to_rank}, GOG~\cite{GOG}, SalMatch~\cite{salicency}, CSBT~\cite{CSBT} and TCP~\cite{DEEP_MULTI_CHANNEL}.
 The detailed comparison results are summarized in Table~\ref{tab:cuhk_Dataset.}. As shown in Table~\ref{tab:cuhk_Dataset.}, the proposed GCT method can achieve competitive results on this dataset. More specifically, we obtain the best rank 1 recognition rate (has a 8.2 percent gain over TCP~\cite{DEEP_MULTI_CHANNEL}, a part based deep learning algorithm). And the recognition rates of other ranks are also competitive.
\begin{table}[!htbp]
\caption{Comparison of top $r$ matching rate using CMC (\%) on CUHK01 dataset. The best and second best results are marked in red and blue, respectively.
\label{tab:cuhk_Dataset.}}
\begin{center}
\begin{tabular}{@{}C{3cm}@{}| @{}C{1.2cm}@{}@{}C{1.2cm}@{}@{}C{1.2cm}@{}@{}C{1.2cm}@{}}
\hline
Methods  & r=1 & r=5 & r=10 & r=20 \\
\hline\hline
Semantic &$32.7$&$51.2$ &$-$&$76.3$ \\
kLFDA &  $32.8$ & $59.0$ & $69.6$ & $-$\\
IDLA &  $47.5$ & $71.5$ & $80.0$ & $-$\\
DeepRanking &  $50.4$ & $75.9$ & $84.1$ & $-$\\
ME &  $53.4$ & $76.3$ & $84.4$ & $-$\\
GOG &  $\textcolor{blue}{\bf57.8}$ & $79.1$ & $86.2$ & $-$\\
SalMatch &$28.5$&$46.0$& $-$&$67.3$\\
CSBT &  $51.2$ & $76.3$ & $-$ & $91.8$\\
TCP &  $53.7$ & $\textcolor{red}{\bf84.3}$ & $\textcolor{red}{\bf91.0}$ & $\textcolor{red}{\bf96.3}$\\
\hline\hline
Our GCT &  $\textcolor{red}{\bf 61.9}$ & $ \textcolor{blue}{\bf81.9}$ & $ \textcolor{blue}{\bf87.6}$ & $ \textcolor{blue}{\bf92.8}$\\
\hline
\end{tabular}
\end{center}
\end{table}
%
%
%
%

\subsection{Analysis on the number of selected references}

Different number of selected references ($R$) for calculating the distances between test pairs have different influences on the re-identification performance. With a small $R$, bad references may have a large impact on the patch-wise distance calculation, resulting in deteriorating the recognition performance. By contrast, if the value of $R$ is large, the correspondences transferred from less similar references may introduce inaccurate correspondences, which also degrades the performance. We record the best $R$ on the VIPeR, Road, PRID450S, 3DPES and CUHK01 are 20, 5, 10, 20 and 20, respectively. As shown in Figure~\ref{fig:knn results}, $R=50$ performs the worst on all the other four datasets (except for VIPeR, on which $R=50$ performs slightly better than $R=5$, but is still inferior than $R=10$ and $R=20$). A relatively small optimal $R$ on Road dataset can be attributed to the similar view of images in this dataset (small $R$ can include enough meaningful patch-wise correspondences for each test pair). While a bigger optimal $R$ on VIPeR, 3DPES and CUHK01 datasets indicates that the pose variations on these datasets are more serious, therefore more references should be incorporated together to include enough correct correspondences.
\subsection{Analysis on Typical Failure Cases}
\begin{figure}[!t]
  \centering
  \includegraphics[width=\linewidth]{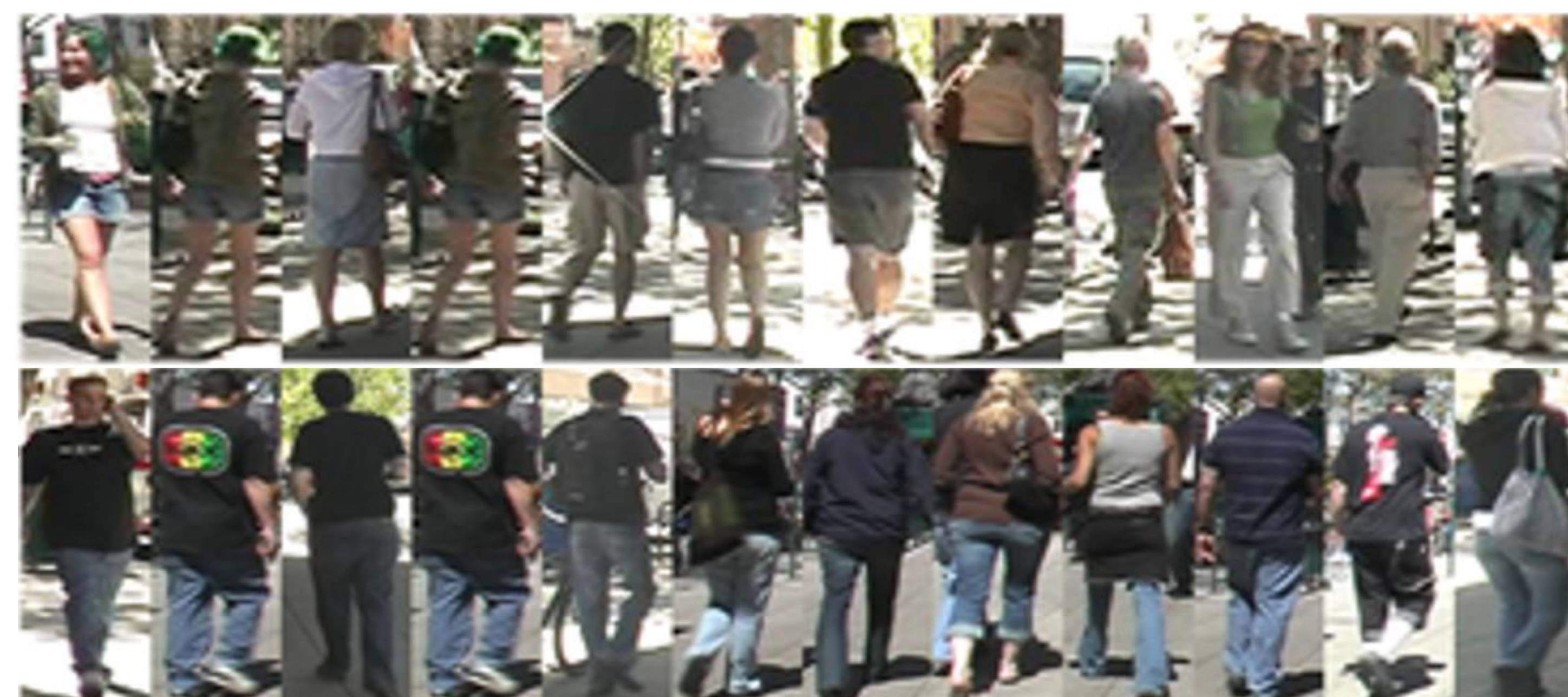}
  \caption{Typical failure cases. The first image in each row is the probe image, and the second one is the correct matched gallery image, the rest images are the ranking list generated by GCT.   }
\label{fig:failure_cases}
\end{figure}

We record the typical failure cases to explore the limitations of the proposed GCT algorithm. As shown in Figure~\ref{fig:failure_cases}, when severe self-occlusion occurs, the appearances of the same person may be dramatically different across different camera views, rendering it difficult for GCT to establish enough local correspondences between matched image pairs. Even though, the proposed GCT algorithm can rank visually similar images with the probe ahead of others, which is valuable for further manual verification.
\section{Conclusion}
\label{sec:conclusion}

This paper proposes a novel GCT model for Re-ID task. The GCT model aims to learn a set of patch-wise correspondence templates from positive image pairs in the training set, and then transfer these correspondences to test image pairs with similar pose-pair configurations for distance computation. Owing to the part-based strategy as well as the incorporation of the body context information, the GCT model is capable of dealing with the problem of spatial misalignment caused by large variations in viewpoints and human poses. Extensive experiments on five challenging datasets demonstrate the effectiveness of the GCT model.\\

\noindent
{\bf Acknowledgments} This work was supported by the National Natural Science Foundation of China (NSFC) (Grant No. 61671289, 61221001, 61771303 and 61571261), and Ling was supported in part by the US NSF (Grant No. 1618398, 1449860, and 1350521).
\bibliographystyle{aaai}\normalsize
\bibliography{reference}

\begin{thebibliography}{}

\bibitem[\protect\citeauthoryear{Ahmed, Jones, and Marks}{2015}]{AhmedJM15}
Ahmed, E.; Jones, M.~J.; and Marks, T.~K.
\newblock 2015.
\newblock An improved deep learning architecture for person re-identification.
\newblock In {\em CVPR}.

\bibitem[\protect\citeauthoryear{Andriluka, Roth, and
  Schiele}{2010}]{TUD_DATASET}
Andriluka, M.; Roth, S.; and Schiele, B.
\newblock 2010.
\newblock Monocular 3d pose estimation and tracking by detection.
\newblock In {\em CVPR}.

\bibitem[\protect\citeauthoryear{Baltieri, Vezzani, and
  Cucchiara}{2011}]{three_dpes}
Baltieri, D.; Vezzani, R.; and Cucchiara, R.
\newblock 2011.
\newblock 3dpes: 3d people dataset for surveillance and forensics.
\newblock In {\em Joint ACM Workshop on Human Gesture and Behavior
  Understanding}.

\bibitem[\protect\citeauthoryear{Chen \bgroup et al\mbox.\egroup
  }{2015}]{ChenYHZW15}
Chen, D.; Yuan, Z.; Hua, G.; Zheng, N.; and Wang, J.
\newblock 2015.
\newblock Similarity learning on an explicit polynomial kernel feature map for
  person re-identification.
\newblock In {\em CVPR}.

\bibitem[\protect\citeauthoryear{Chen \bgroup et al\mbox.\egroup
  }{2016}]{sim_spatial_constraints}
Chen, D.; Yuan, Z.; Chen, B.; and Zheng, N.
\newblock 2016.
\newblock Similarity learning with spatial constraints for person
  re-identification.
\newblock In {\em CVPR}.

\bibitem[\protect\citeauthoryear{Chen \bgroup et al\mbox.\egroup }{2017}]{CSBT}
Chen, J.; Wang, Y.; Qin, J.; Liu, L.; and Shao, L.
\newblock 2017.
\newblock Fast person re-identification via cross-camera semantic binary
  transformation.
\newblock In {\em CVPR}.

\bibitem[\protect\citeauthoryear{Chen, Guo, and Lai}{2016}]{deep_ranking}
Chen, S.; Guo, C.; and Lai, J.
\newblock 2016.
\newblock Deep ranking for person re-identification via joint representation
  learning.
\newblock {\em IEEE Transactions. Image Processing} 25(5):2353--2367.

\bibitem[\protect\citeauthoryear{Cheng \bgroup et al\mbox.\egroup
  }{2016}]{DEEP_MULTI_CHANNEL}
Cheng, D.; Gong, Y.; Zhou, S.; Wang, J.; and Zheng, N.
\newblock 2016.
\newblock Person re-identification by multi-channel parts-based {CNN} with
  improved triplet loss function.
\newblock In {\em CVPR}.

\bibitem[\protect\citeauthoryear{Farenzena \bgroup et al\mbox.\egroup
  }{2010}]{FarenzenaBPMC10}
Farenzena, M.; Bazzani, L.; Perina, A.; Murino, V.; and Cristani, M.
\newblock 2010.
\newblock Person re-identification by symmetry-driven accumulation of local
  features.
\newblock In {\em CVPR}.

\bibitem[\protect\citeauthoryear{Gray, Brennan, and Tao}{2007}]{viper}
Gray, D.; Brennan, S.; and Tao, H.
\newblock 2007.
\newblock Evaluating appearance models for recognition, reacquisition, and
  tracking.
\newblock In {\em PETS Workshop}.

\bibitem[\protect\citeauthoryear{Jose and Fleuret}{2016}]{WARCA}
Jose, C., and Fleuret, F.
\newblock 2016.
\newblock Scalable metric learning via weighted approximate rank component
  analysis.
\newblock In {\em ECCV}.

\bibitem[\protect\citeauthoryear{Karanam, Li, and Radke}{2015}]{KaranamLR15}
Karanam, S.; Li, Y.; and Radke, R.~J.
\newblock 2015.
\newblock Person re-identification with discriminatively trained viewpoint
  invariant dictionaries.
\newblock In {\em ICCV}.

\bibitem[\protect\citeauthoryear{K{\"{o}}stinger \bgroup et al\mbox.\egroup
  }{2012}]{KISSME}
K{\"{o}}stinger, M.; Hirzer, M.; Wohlhart, P.; Roth, P.~M.; and Bischof, H.
\newblock 2012.
\newblock Large scale metric learning from equivalence constraints.
\newblock In {\em CVPR}.

\bibitem[\protect\citeauthoryear{Li \bgroup et al\mbox.\egroup }{2013}]{SVMML}
Li, Z.; Chang, S.; Liang, F.; Huang, T.~S.; Cao, L.; and Smith, J.~R.
\newblock 2013.
\newblock Learning locally-adaptive decision functions for person verification.
\newblock In {\em CVPR}.

\bibitem[\protect\citeauthoryear{Li, Zhao, and Wang}{2012}]{CUHK01}
Li, W.; Zhao, R.; and Wang, X.
\newblock 2012.
\newblock Human reidentification with transferred metric learning.
\newblock In {\em ACCV}.

\bibitem[\protect\citeauthoryear{Liao \bgroup et al\mbox.\egroup }{2015}]{lomo}
Liao, S.; Hu, Y.; Zhu, X.; and Li, S.~Z.
\newblock 2015.
\newblock Person re-identification by local maximal occurrence representation
  and metric learning.
\newblock In {\em CVPR}.

\bibitem[\protect\citeauthoryear{Liaw and Wiener}{2002}]{random_forest}
Liaw, A., and Wiener, M.
\newblock 2002.
\newblock Classification and regression by random forest.
\newblock {\em R News}.

\bibitem[\protect\citeauthoryear{Martinel \bgroup et al\mbox.\egroup
  }{2016}]{TMA}
Martinel, N.; Das, A.; Micheloni, C.; and Roy{-}Chowdhury, A.~K.
\newblock 2016.
\newblock Temporal model adaptation for person re-identification.
\newblock In {\em ECCV}.

\bibitem[\protect\citeauthoryear{Matsukawa \bgroup et al\mbox.\egroup
  }{2016}]{GOG}
Matsukawa, T.; Okabe, T.; Suzuki, E.; and Sato, Y.
\newblock 2016.
\newblock Hierarchical gaussian descriptor for person re-identification.
\newblock In {\em CVPR}.

\bibitem[\protect\citeauthoryear{Mignon and Jurie}{2012}]{pcca}
Mignon, A., and Jurie, F.
\newblock 2012.
\newblock {PCCA:} {A} new approach for distance learning from sparse pairwise
  constraints.
\newblock In {\em CVPR}.

\bibitem[\protect\citeauthoryear{Oreifej, Mehran, and Shah}{2010}]{OreifejMS10}
Oreifej, O.; Mehran, R.; and Shah, M.
\newblock 2010.
\newblock Human identity recognition in aerial images.
\newblock In {\em CVPR}.

\bibitem[\protect\citeauthoryear{Paisitkriangkrai, Shen, and van~den
  Hengel}{2015}]{learn_to_rank}
Paisitkriangkrai, S.; Shen, C.; and van~den Hengel, A.
\newblock 2015.
\newblock Learning to rank in person re-identification with metric ensembles.
\newblock In {\em CVPR}.

\bibitem[\protect\citeauthoryear{Pedagadi \bgroup et al\mbox.\egroup
  }{2013}]{lfda}
Pedagadi, S.; Orwell, J.; Velastin, S.~A.; and Boghossian, B.~A.
\newblock 2013.
\newblock Local fisher discriminant analysis for pedestrian re-identification.
\newblock In {\em CVPR}.

\bibitem[\protect\citeauthoryear{Roth \bgroup et al\mbox.\egroup
  }{2014}]{prid_450s}
Roth, P.~M.; Hirzer, M.; K{\"{o}}stinger, M.; Beleznai, C.; and Bischof, H.
\newblock 2014.
\newblock Mahalanobis distance learning for person re-identification.
\newblock In {\em Person Re-Identification}.

\bibitem[\protect\citeauthoryear{Shen \bgroup et al\mbox.\egroup
  }{2015}]{iccv15_correspondence}
Shen, Y.; Lin, W.; Yan, J.; Xu, M.; Wu, J.; and Wang, J.
\newblock 2015.
\newblock Person re-identification with correspondence structure learning.
\newblock In {\em ICCV}.

\bibitem[\protect\citeauthoryear{Shi, Hospedales, and Xiang}{2015}]{Semantic15}
Shi, Z.; Hospedales, T.~M.; and Xiang, T.
\newblock 2015.
\newblock Transferring a semantic representation for person re-identification
  and search.
\newblock In {\em CVPR}.

\bibitem[\protect\citeauthoryear{Suh, Adamczewski, and Lee}{2015}]{SuhAL15}
Suh, Y.; Adamczewski, K.; and Lee, K.~M.
\newblock 2015.
\newblock Subgraph matching using compactness prior for robust feature
  correspondence.
\newblock In {\em CVPR}.

\bibitem[\protect\citeauthoryear{Xiao \bgroup et al\mbox.\egroup }{2016}]{DGD}
Xiao, T.; Li, H.; Ouyang, W.; and Wang, X.
\newblock 2016.
\newblock Learning deep feature representations with domain guided dropout for
  person re-identification.
\newblock In {\em CVPR}.

\bibitem[\protect\citeauthoryear{Xiong \bgroup et al\mbox.\egroup
  }{2014}]{kLFDA14}
Xiong, F.; Gou, M.; Camps, O.~I.; and Sznaier, M.
\newblock 2014.
\newblock Person re-identification using kernel-based metric learning methods.
\newblock In {\em ECCV}.

\bibitem[\protect\citeauthoryear{Yang \bgroup et al\mbox.\egroup
  }{2014}]{YangYYLYL14}
Yang, Y.; Yang, J.; Yan, J.; Liao, S.; Yi, D.; and Li, S.~Z.
\newblock 2014.
\newblock Salient color names for person re-identification.
\newblock In {\em ECCV}.

\bibitem[\protect\citeauthoryear{Yang \bgroup et al\mbox.\egroup
  }{2017}]{YangWLL17}
Yang, Y.; Wen, L.; Lyu, S.; and Li, S.~Z.
\newblock 2017.
\newblock Unsupervised learning of multi-level descriptors for person
  re-identification.
\newblock In {\em AAAI}.

\bibitem[\protect\citeauthoryear{Zhang \bgroup et al\mbox.\egroup
  }{2016a}]{DCSL}
Zhang, Y.; Li, X.; Zhao, L.; and Zhang, Z.
\newblock 2016a.
\newblock Semantics-aware deep correspondence structure learning for robust
  person re-identification.
\newblock In {\em IJCAI}.

\bibitem[\protect\citeauthoryear{Zhang \bgroup et al\mbox.\egroup
  }{2016b}]{LSSCDL16}
Zhang, Y.; Li, B.; Lu, H.; Irie, A.; and Ruan, X.
\newblock 2016b.
\newblock Sample-specific {SVM} learning for person re-identification.
\newblock In {\em CVPR}.

\bibitem[\protect\citeauthoryear{Zhang, Xiang, and Gong}{2016}]{NSFT}
Zhang, L.; Xiang, T.; and Gong, S.
\newblock 2016.
\newblock Learning a discriminative null space for person re-identification.
\newblock In {\em CVPR}.

\bibitem[\protect\citeauthoryear{Zhao \bgroup et al\mbox.\egroup
  }{2017}]{Spindle_Net}
Zhao, H.; Tian, M.; Sun, S.; Shao, J.; Yan, J.; Yi, S.; Wng, X.; and Tang, X.
\newblock 2017.
\newblock Spindle net: Person re-identification with human body region guided
  feature decomposition and fusion.
\newblock In {\em CVPR}.

\bibitem[\protect\citeauthoryear{Zhao, Ouyang, and Wang}{2013}]{salicency}
Zhao, R.; Ouyang, W.; and Wang, X.
\newblock 2013.
\newblock Unsupervised salience learning for person re-identification.
\newblock In {\em CVPR}.

\bibitem[\protect\citeauthoryear{Zhao, Ouyang, and Wang}{2014}]{mid_filter14}
Zhao, R.; Ouyang, W.; and Wang, X.
\newblock 2014.
\newblock Learning mid-level filters for person re-identification.
\newblock In {\em CVPR}.

\bibitem[\protect\citeauthoryear{Zheng \bgroup et al\mbox.\egroup
  }{2015}]{ZhengSTWWT15}
Zheng, L.; Shen, L.; Tian, L.; Wang, S.; Wang, J.; and Tian, Q.
\newblock 2015.
\newblock Scalable person re-identification: {A} benchmark.
\newblock In {\em ICCV}.

\end{thebibliography}

\end{document}